\pdfoutput=1
\documentclass[letterpaper]{article}
\usepackage[preprint]{aaai2027}
\usepackage[hyphens]{url}
\usepackage{graphicx}
\usepackage{natbib}
\usepackage{caption}
\usepackage{amsmath,amssymb,mathtools}
\usepackage{booktabs}
\usepackage{multirow}
\usepackage{array}
\usepackage{xspace}
\usepackage{pifont}

\newcommand{\method}{SeGDeP\xspace}
\newcommand{\samthree}{SAM~3\xspace}

\newcommand{\finalanswer}{\texttt{<final\_answer>}}

\title{SeGDeP: Semantic- and Geometric-Aware Decoupled Prompts for Reasoning Segmentation}
\author{
  Linnan Zhao,
  Xu Liu,
  Lingling Li,\\
  Licheng Jiao,
  Fang Liu,
  Wenping Ma
}
\affiliations{
  Xidian University\\
  Xi'an, China
}

\begin{document}
\maketitle

\begin{abstract}
Reasoning segmentation converts an implicit linguistic conclusion into a precise mask, requiring both semantic identification and spatial grounding. Existing MLLM--segmenter interfaces either use a special trigger or compress both signals into one context, although they receive different supervision and fail differently. This coupling obscures whether a failure arises from target interpretation or from localization. We present \method, an \textbf{explicit \emph{what--where} interface}. A semantic prompt branch and an independent geometric projection path transform resolved MLLM states into semantic features and a DETR-predicted box, which jointly condition a SAM~3 mask decoder. Training first aligns this executable interface, then uses group reward-decoupled policy optimization (GDPO) to balance format, box-IoU, and mask-IoU feedback. \method-4B reaches \textbf{82.7 average cIoU} over eight RefCOCO-family splits and 66.0/59.6 gIoU on ReasonSeg val/test while adapting \textbf{only 0.38\%} of Qwen3-VL parameters through LoRA. Controlled stage-wise ablations, gradient diagnostics, and prompt interventions further show that the two paths develop complementary semantic and geometric specialization rather than duplicating the same evidence.
\end{abstract}

\section{Introduction}
Reasoning segmentation resolves an object described indirectly by function, relation, or likely action, then grounds it as a pixel mask \cite{rssurvey}. It couples \emph{what} satisfies the instruction with \emph{where} it lies: the former requires categorical, attribute, and relational evidence, whereas the latter requires localization and boundary recovery. MLLMs favor semantic inference \cite{llava,instructblip,gemini,qwen2vl,qwen25vl,qwen3vl}, and promptable segmenters geometric execution \cite{sam,sam2,sam3}; their interface is therefore central. A useful interface must expose both signals in forms that the downstream segmenter can execute and supervise. This distinction becomes critical when visually similar instances satisfy only part of an indirect description. A single opaque context also obscures whether an error arose from interpreting the referent or locating it.

\begin{figure}[t]
  \centering
  \includegraphics[width=\columnwidth]{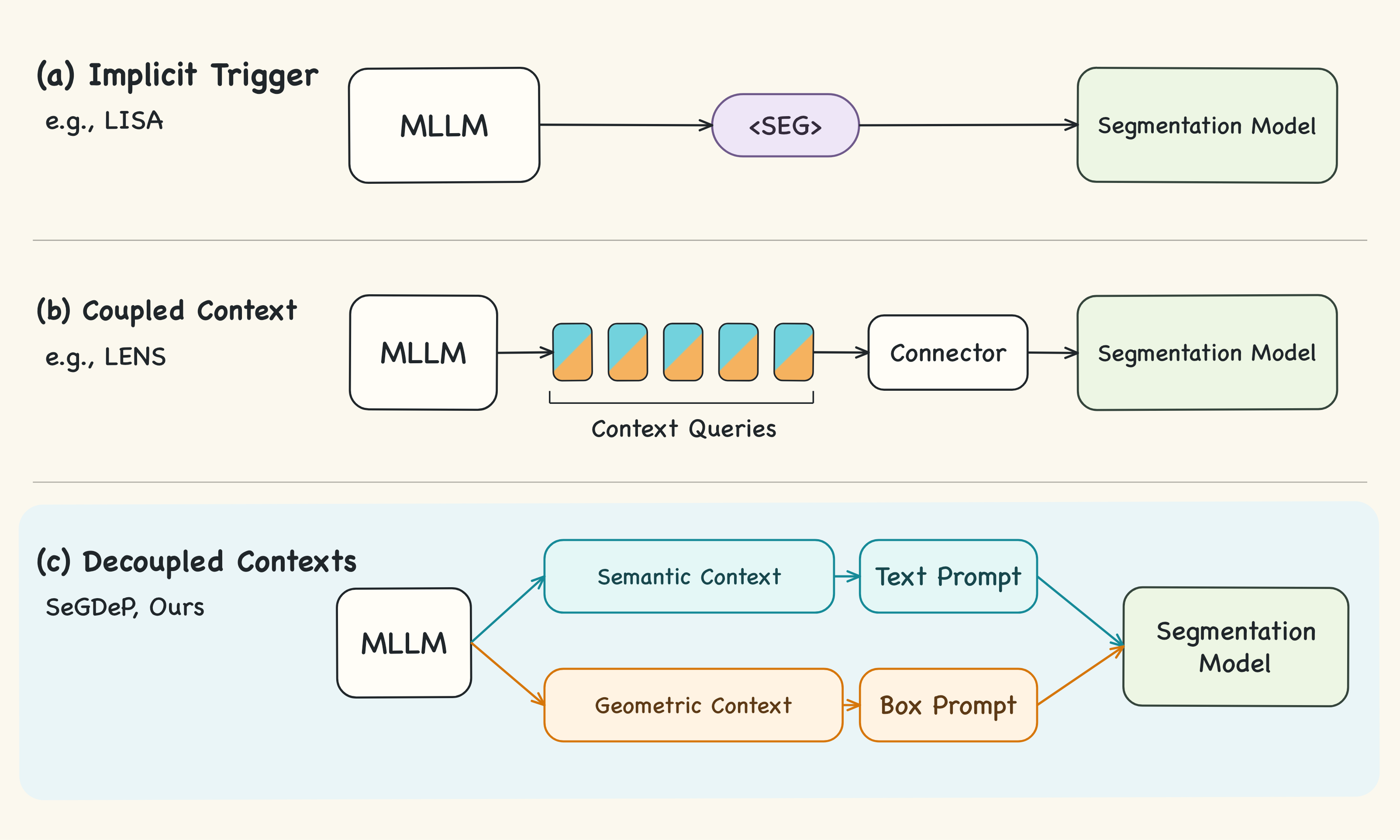}
  \caption{Reasoning-to-segmentation interfaces. Trigger and coupled-context designs provide no explicit separation; \method emits complementary semantic and box prompts.}
  \label{fig:paradigms}
\end{figure}

Figure~\ref{fig:paradigms} contrasts implicit-trigger and coupled-context interfaces with our explicit decoupled design for segmentation. LISA-like systems use a special \texttt{<SEG>} token as an implicit trigger \cite{lisa,lisaplus}; LENS pools the reasoning trace into a richer context \cite{lens}. Both require one latent stream to preserve identity and regress coordinates. Their heterogeneous supervision creates asymmetric errors: the correct role but wrong instance, or a plausible region obtained by misreading the target relation. In a controlled shared-context model, 37--46\% of samples yield opposing semantic and geometric gradients, exposing data-induced demands that often compete at one bottleneck.

We introduce \method, which makes both decisions explicit and executable. A semantic context extractor produces role-typed prompt features, while an independent geometric projection preserves context tokens for localization. They jointly condition DETR, frozen image features refine its coarse box, and the mask decoder combines the resulting semantic and box prompts. Thus, prompt construction is separated before downstream integration and supervised in each execution space.

Training first aligns the interface, then optimizes reasoning with format, box-IoU, and mask-IoU feedback normalized by GDPO \cite{gdpo}. Matched coupled baselines, stage-wise checkpoints, gradient measurements, and prompt interventions test the mechanism. \method reaches 82.7 average cIoU across eight RefCOCO-family splits and 66.0/59.6 gIoU on ReasonSeg val/test.

Our main contributions are:
\begin{itemize}
  \item An executable semantic--geometric interface converts resolved MLLM states into specialized text and box prompts through semantic prompt refinement and geometric context projection.
  \item Comprehensive ablations verify the design, while gradient diagnostics and prompt interventions expose complementary semantic and geometric roles.
  \item Interface alignment and GDPO deliver strong RefCOCO and ReasonSeg results while adapting only 0.38\% of Qwen3-VL parameters through LoRA.
\end{itemize}

\section{Related Work}
\paragraph{Image reasoning segmentation.}
Referring image segmentation grounds explicit expressions at pixel level \cite{refcoco,refcocoplus,referit}. Methods such as LAVT and ReLA improve cross-modal alignment \cite{lavt,gres}, while PixelLM, GLaMM, SAM4MLLM, and UniPixel connect MLLMs to pixel decoders \cite{pixellm,glamm,sam4mllm,unipixel}. Reasoning segmentation extends this setting to implicit attributes, functions, and relations. LISA and LISA++ use a segmentation token, Seg-Zero emits spatial prompts, ThinkFirst structures rationales, and LENS pools chain-of-thought states \cite{lisa,lisaplus,segzero,thinkfirst,lens}. These interfaces improve reasoning, but still encode semantic identity and spatial support in a single implicit or coupled representation.

Promptable segmenters expose executable controls: SAM and SAM~2 accept spatial prompts, and \samthree adds concept-aware text conditioning \cite{sam,sam2,sam3}; grounding models likewise demonstrate structured language alignment with box regression \cite{groundingdino,glip}. \method therefore translates resolved reasoning into separate text and box prompts, retaining semantic discrimination while making localization explicit and measurable.

\paragraph{Chain-of-thought and policy optimization.}
Chain-of-thought and its multimodal extensions externalize intermediate decisions and visual evidence \cite{cot,zeroshotcot,mmcot,vcot}. ThinkFirst structures rationales for reasoning segmentation \cite{thinkfirst}; we use a Self-Ask-style QA trace to make intermediate conclusions parseable \cite{measuring}.
For dense prediction, a well-formed rationale may still select the wrong instance or yield an unusable prompt.

GRPO uses within-group relative outcomes and requires no learned critic \cite{grpo}; Seg-Zero and LENS extend it to segmentation-aware reasoning. GDPO separately normalizes format, localization, and mask rewards before aggregation because their scales differ \cite{gdpo}.

\section{Method}
\label{sec:method}

\begin{figure*}[t]
  \centering
  \includegraphics[width=\textwidth]{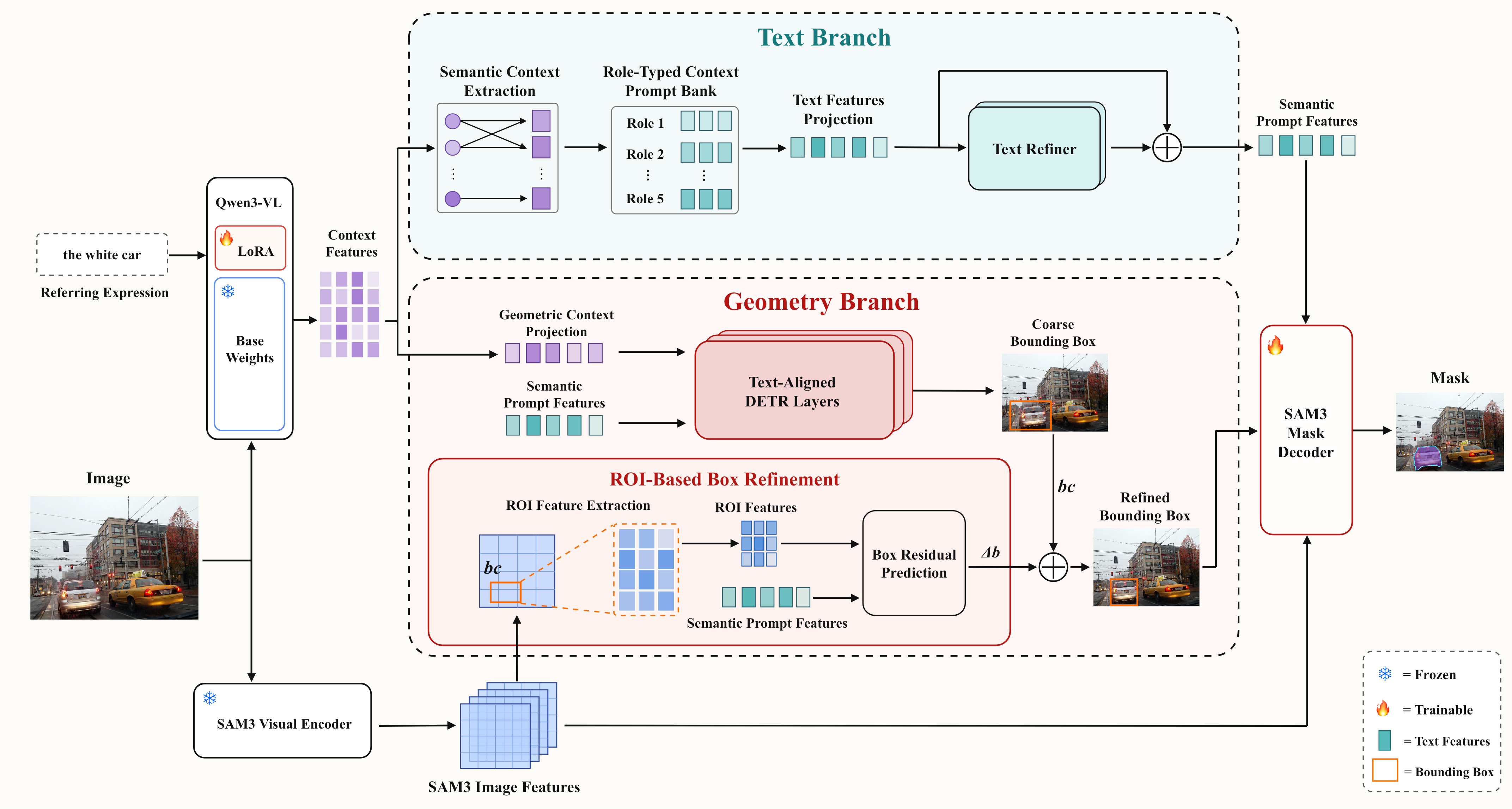}
  \caption{Architecture of the Qwen3-VL/\samthree instantiation of \method. A role-typed semantic branch produces $P_{\mathrm{sem}}$, while an independent projection produces geometric context $Z_{\mathrm{geo}}$. Both condition DETR to predict $b_c$, which selects frozen \samthree ROI features for residual refinement. The decoder combines semantic features, the refined box, and image features. Snowflakes and flames denote frozen base weights and trainable adapters or decoder components, respectively; all newly introduced modules are trainable.}
  \label{fig:architecture}
\end{figure*}

\subsection{Problem Formulation and Overview}
Let $x=(I,q)$ denote an image and a referring request, either explicit or implicit; the goal is to predict its binary mask $M^*$. An MLLM produces a QA trace $y$ and hidden states $H$, while the frozen \samthree image backbone extracts $F$. A semantic context-query extractor attends to $H$, and an independent projection retains the hidden-state sequence for geometric decoding:
\begin{equation}
  C_{\mathrm{sem}}=\operatorname{Attn}(Q_{\mathrm{sem}},H,H),
  \qquad Z_{\mathrm{geo}}=\Pi_{\mathrm{geo}}(H).
  \label{eq:contexts}
\end{equation}
$C_{\mathrm{sem}}$ represents the identity, attributes, and relations needed to resolve \emph{what}, while $Z_{\mathrm{geo}}$ retains the multimodal evidence needed to determine \emph{where}. The two parameterized paths specialize before integration.

The semantic branch produces $P_{\mathrm{sem}}=\mathcal T(C_{\mathrm{sem}})$, while the geometric path predicts $b=\mathcal G(F,Z_{\mathrm{geo}},P_{\mathrm{sem}})$. Their joint decoding is
\begin{equation}
  \hat M=\sigma\!\left(\mathcal D(F,P_{\mathrm{sem}},b)\right).
  \label{eq:overview}
\end{equation}
Here $\mathcal T$ translates semantic context into the \samthree prompt space, $\mathcal G$ denotes coarse-to-fine localization, and $\mathcal D$ is the mask decoder. The paths specialize before the decoder recombines identity and spatial support.
$F$ is shared by ROI refinement and mask decoding. $P_{\mathrm{sem}}$ enters both the DETR memory and the native text-prompt channel, whereas $b$ is supplied through the spatial-prompt channel.

\subsection{Decoupled Prompt Connector}
\paragraph{Semantic prompt: what to segment.}
The translator organizes $C_{\mathrm{sem}}$ with a role-typed bank $\mathcal B$, projects its tokens into the \samthree text space, and applies a residual refiner: $P_0=\Pi_{\mathrm{text}}(\mathcal B(C_{\mathrm{sem}}))$ and $P_{\mathrm{sem}}=P_0+\mathcal R_{\mathrm{text}}(P_0)$. The bank maps the reasoning context to a fixed prompt-token set whose learnable roles capture complementary attributes and relations. $\Pi_{\mathrm{text}}$ matches the \samthree prompt representation, and the residual update refines the token content while preserving its initial projection.

\paragraph{Geometric prompt: coarse localization.}
The geometric projection in Eq.~\ref{eq:contexts} maps the MLLM hidden-state sequence directly into localization tokens $Z_{\mathrm{geo}}$, preserving its sequence for spatial selection by DETR. We concatenate it with the semantic prompt to form the text-aligned memory $U_{\mathrm{geo}}=[Z_{\mathrm{geo}};P_{\mathrm{sem}}]$. Using one referent query, the decoder produces a normalized coarse box
\begin{equation}
  b_c=\mathcal G_c(U_{\mathrm{geo}}).
  \label{eq:coarse_box}
\end{equation}
$Z_{\mathrm{geo}}$ carries MLLM spatial evidence, while $P_{\mathrm{sem}}$ aligns the decoder with the resolved referent. Successive decoder layers refine the single-referent estimate, and the final $b_c$ enters ROI refinement.
At each layer, the referent query cross-attends to $U_{\mathrm{geo}}$ and updates its reference box.

\paragraph{ROI box refinement.}
The coarse box selects local features $R=\operatorname{ROIAlign}(F,b_c)$. The ROI refiner combines these boundary-sensitive features with the semantic prompt and predicts the final box as
\begin{equation}
  b=b_c+\mathcal R(R,P_{\mathrm{sem}},b_c).
  \label{eq:roi_refine}
\end{equation}
Here $\mathcal R$ predicts a residual correction. Local visual evidence improves boundary alignment, while $P_{\mathrm{sem}}$ distinguishes nearby instances with similar spatial support. The two sources are fused during box refinement, while the upstream semantic and geometric paths remain separately parameterized.
The coarse box initializes the refinement query, and the ROI tokens and $P_{\mathrm{sem}}$ form its local visual--semantic memory.

\paragraph{Composite mask decoding.}
The refined box and semantic tokens enter the native spatial- and text-prompt channels of the \samthree mask decoder, which jointly conditions on the image features and both prompts. The box restricts the spatial support, whereas the semantic prompt preserves the identity cues needed to distinguish overlapping or visually similar instances. Mask gradients reach both prompt paths, which remain separately parameterized until their downstream integration.
The \samthree prompt transformer fuses these inputs before the segmentation head produces mask logits.

\subsection{Two-Stage Optimization}
The two stages separate learning an executable prompt interface from adapting the reasoning policy that drives it.

\paragraph{Stage 1: interface alignment.}
We freeze the MLLM, the \samthree image backbone, and its mask decoder, and optimize the two prompt paths and coarse-to-fine localization modules in Fig.~\ref{fig:architecture}. A frozen \samthree text encoder maps the expression to teacher prompt $P_{\mathrm{sem}}^*$, while $b^*$ is derived from the target mask. Let $\mathcal V$ index the valid semantic-prompt tokens. The objective is
\begin{equation}
  \begin{aligned}
  \mathcal L_1={}&\frac{\lambda_s}{|\mathcal V|}
  \sum_{t\in\mathcal V}
  \|p_{\mathrm{sem},t}-p_{\mathrm{sem},t}^*\|_2^2
  +\lambda_b\mathcal L_{\rm box},\\
  \mathcal L_{\rm box}={}&\|b-b^*\|_1
  +\mathcal L_{\rm GIoU}(b,b^*).
  \end{aligned}
  \label{eq:stage1}
\end{equation}
The semantic term aligns the learned prompt with the \samthree text-prompt space, while the box terms supervise geometric localization. Stage~1 establishes a stable mapping from MLLM hidden states to executable text and box prompts, providing the initialization for policy optimization.
The frozen text encoder supplies training targets, and inference uses the learned semantic path.

\paragraph{Stage 2: reasoning elicitation with GDPO.}
We activate LoRA in the MLLM \cite{lora}, keep the image backbone frozen, and update the connector and mask decoder. For each input, we sample $G$ reasoning traces. Completion $i$ receives mask reward $R_{m,i}=\operatorname{IoU}(\hat M_i,M^*)$, box reward $R_{b,i}=\operatorname{IoU}(b_i,b^*)$, and format reward $R_{f,i}$. These components score final mask quality, spatial localization, and structured reasoning with an explicit referent, respectively.

These rewards differ in scale and variance. GDPO \cite{gdpo} standardizes each component within the completions sampled for the same input before aggregation:
\begin{equation}
  A_i=\sum_{j\in\{m,b,f\}}w_j
  \frac{R_{j,i}-\mu_g(R_j)}
       {\sigma_g(R_j)+\epsilon}.
  \label{eq:gdpo}
\end{equation}
Component-wise standardization balances the contributions of format, localization, and mask quality to the group advantage.

The final objective combines the advantage-weighted policy loss with a KL constraint and supervised geometric and mask stabilization:
\begin{equation}
  \mathcal L_2=\mathcal L_{\rm PG}(A)
  +\beta\mathcal L_{\rm KL}(\pi_\theta,\pi_{\rm ref})
  +\lambda_b\mathcal L_{\rm box}
  +\lambda_m\mathcal L_{\rm mask}.
  \label{eq:stage2}
\end{equation}
Here $\mathcal L_{\rm PG}$ increases the likelihood of sampled traces with positive $A_i$ and suppresses those with negative $A_i$. The reference $\pi_{\rm ref}$ is the frozen policy at the start of Stage~2, and $\mathcal L_{\rm mask}=\mathcal L_{\rm Dice}+\mathcal L_{\rm BCE}$ \cite{dice}. The box objective from Eq.~\ref{eq:stage1} \cite{giou} and mask supervision preserve the executable interface during policy optimization. Stage~2 jointly updates the prompt translator, localization path, and mask decoder under this objective. Exact format scoring, sample-equal completion-token averaging, and the $k_3$ estimator \cite{k3} used for $\mathcal L_{\rm KL}$ are detailed in Appendix~\ref{app:gdpo}.

\section{Experiments}
\subsection{Setup}
\paragraph{Datasets and metrics.}
We evaluate explicit referring segmentation on RefCOCO and RefCOCO+ \cite{refcocoplus}, and RefCOCOg \cite{refcoco}, and implicit reasoning segmentation on ReasonSeg \cite{lisa}, using their official splits. GroundingSuite-Eval (GSEval) \cite{groundingsuite} is held out for zero-shot transfer. Cumulative IoU (cIoU) pools intersections and unions over a split and therefore gives larger masks more weight; generalized IoU (gIoU) averages per-image mask IoU. We follow the dominant protocol by using cIoU for the RefCOCO-series comparison and both metrics on ReasonSeg.

\paragraph{Models and comparison protocol.}
The main model pairs Qwen3-VL-4B-Instruct with \samthree. Stage~1 trains the executable interface on the RefCOCO series; Stage~2 adds ReasonSeg, continues updating the connector, and activates 17M MLLM LoRA parameters (0.38\% of Qwen3-VL) and the \samthree mask decoder while keeping the base MLLM and image backbone frozen. Both stages use AdamW \cite{adamw}. GSEval is used for neither training nor model selection. An additional 2B variant replaces only the MLLM with Qwen3-VL-2B-Instruct and otherwise retains the segmenter, data, optimization, and evaluation protocol. Table~\ref{tab:refcoco} separates methods with and without active chain-of-thought reasoning, while Table~\ref{tab:reasonseg} reports results on ReasonSeg. Prior-method scores are taken from their reported official-split results. Complete hyperparameters and trainable scopes are provided in Appendix~\ref{app:implementation}.

\subsection{Main Results}
\begin{table*}[t]
\centering
{\small
\setlength{\tabcolsep}{1.2pt}
\begin{tabular*}{\textwidth}{@{\extracolsep{\fill}}llccccccccc@{}}
\toprule
& & \multicolumn{3}{c}{RefCOCO} & \multicolumn{3}{c}{RefCOCO+} & \multicolumn{2}{c}{RefCOCOg} & \\
Method & Venue & val & testA & testB & val & testA & testB & val-u & test-u & Avg.\\
\midrule
\multicolumn{11}{l}{\emph{Without active CoT reasoning}}\\
LAVT \cite{lavt} & CVPR'22 & 72.7 & 75.8 & 68.8 & 62.1 & 68.4 & 55.1 & 61.2 & 62.1 & 65.8\\
ReLA \cite{gres} & CVPR'23 & 73.8 & 76.5 & 70.2 & 66.0 & 71.0 & 57.7 & 65.0 & 66.0 & 68.3\\
LISA-7B \cite{lisa} & CVPR'24 & 74.1 & 76.5 & 71.1 & 62.4 & 67.4 & 56.5 & 66.4 & 68.5 & 67.9\\
PixelLM-7B \cite{pixellm} & CVPR'24 & 76.9 & 78.5 & 74.4 & 69.2 & 72.1 & 64.5 & 70.7 & 72.4 & 72.3\\
PerceptionGPT \cite{perceptiongpt} & CVPR'24 & 75.1 & 78.6 & 71.7 & 68.5 & 73.9 & 61.3 & 70.3 & 71.7 & 71.4\\
OMG-LLaVA \cite{omgllava} & NeurIPS'24 & 78.0 & 80.3 & 74.1 & 69.1 & 73.1 & 63.0 & 72.9 & 72.9 & 72.9\\
SAM4MLLM-8B \cite{sam4mllm} & ECCV'24 & 79.8 & 82.7 & 74.7 & 74.6 & 80.0 & 67.2 & 75.5 & 76.4 & 76.4\\
VISA \cite{visa} & ECCV'24 & 72.4 & 75.5 & 68.1 & 59.8 & 64.8 & 53.1 & 65.5 & 66.4 & 65.7\\
GLaMM-7B \cite{glamm} & CVPR'24 & 79.5 & 83.2 & 76.9 & 72.6 & 78.7 & 64.6 & 74.2 & 74.9 & 75.6\\
UniPixel-7B \cite{unipixel} & NeurIPS'25 & 80.8 & 83.0 & 77.4 & 75.3 & 80.1 & 70.0 & 76.4 & 77.1 & 77.5\\
SAM3-Agent-Gemini \cite{sam3} & ICLR'26 & 74.9 & 77.8 & 69.9 & 66.9 & 71.1 & 62.4 & 73.3 & 73.6 & 71.2\\
\midrule
\multicolumn{11}{l}{\emph{With active CoT reasoning}}\\
Seg-Zero-3B \cite{segzero} & arXiv'25 & -- & 79.3 & -- & -- & 73.7 & -- & 71.5 & -- & --\\
Seg-Zero-7B \cite{segzero} & arXiv'25 & -- & 80.3 & -- & -- & 76.2 & -- & 72.6 & -- & --\\
LENS-2B \cite{lens} & AAAI'26 & 80.7 & 82.7 & 77.1 & 73.8 & 78.0 & 67.3 & 75.7 & 76.4 & 76.5\\
LENS-3B \cite{lens} & AAAI'26 & \underline{84.2} & \underline{85.3} & \underline{81.0} & \underline{79.4} & \underline{82.8} & \underline{74.3} & \textbf{81.2} & \underline{81.0} & \underline{81.2}\\
\textbf{\method-2B} & -- & 81.4 & 83.1 & 77.9 & 75.8 & 80.3 & 71.6 & 75.3 & 77.0 & 77.8\\
\textbf{\method-4B} & -- & \textbf{84.2} & \textbf{85.7} & \textbf{82.8} & \textbf{82.3} & \textbf{84.7} & \textbf{79.6} & \underline{80.1} & \textbf{81.9} & \textbf{82.7}\\
\bottomrule
\end{tabular*}
}
\caption{cIoU on the RefCOCO series. Avg. is the mean over eight splits and is unavailable for methods reporting only selected splits. Best and second-best complete results are marked in bold and underline; ranks are determined before one-decimal rounding.}
\label{tab:refcoco}
\end{table*}

\paragraph{RefCOCO series.}
Table~\ref{tab:refcoco} reports the comparison over eight official splits. \method-4B obtains the best average cIoU of 82.7, exceeding LENS-3B by 1.5 points and ranking first on seven splits. The corresponding average per-image gIoU of \method-4B is 83.2, closely tracking its 82.7 cIoU. The largest cIoU margin is +5.3 on RefCOCO+ testB. Because RefCOCO+ removes absolute-location words, its expressions place greater weight on attributes, relations, and instance identity. Averaged over its three splits, \method improves over LENS by 3.4 points, compared with 0.7 on RefCOCO. The larger gain under weaker spatial-language cues highlights the contribution of semantic prompting to instance discrimination.

The compact models in the same CoT group show the same pattern. \method-2B outperforms Seg-Zero-3B on all three commonly reported splits, averages 77.8 versus 76.5 for LENS-2B, and leads LENS on seven of eight splits. Its average gain is 2.9 points on RefCOCO+ versus 0.6 on RefCOCO, reaching +4.3 on RefCOCO+ testB. This cross-scale consistency shows that the benefit of the interface persists at smaller MLLM capacity. The context-plus-connector interface also uses 318.5M parameters versus 474.8M for LENS, a 32.9\% reduction, while Stage~2 adapts 17M MLLM parameters through LoRA (Table~\ref{tab:app_latency}). Higher accuracy with a lighter interface and limited MLLM adaptation further demonstrates the effectiveness of semantic--geometric prompt translation.

\begin{table}[t]
\centering
{\small
\setlength{\tabcolsep}{3.4pt}
\begin{tabular*}{\columnwidth}{@{\extracolsep{\fill}}lcccc@{}}
\toprule
& \multicolumn{2}{c}{Val} & \multicolumn{2}{c}{Test}\\
Method & gIoU & cIoU & gIoU & cIoU\\
\midrule
SAM4MLLM-8B \cite{sam4mllm} & 58.4 & 60.4 & -- & --\\
HyperSeg-3B \cite{hyperseg} & 59.2 & 56.7 & -- & --\\
InstructSeg-3B \cite{instructseg} & 61.9 & \underline{65.2} & -- & --\\
LISA-7B \cite{lisa} & 52.9 & 54.0 & 55.6 & 56.9\\
Seg-Zero-3B \cite{segzero} & 58.2 & 53.1 & 56.1 & 48.6\\
LENS-3B \cite{lens} & \underline{62.1} & 64.9 & \underline{57.2} & \underline{58.0}\\
\textbf{\method-4B} & \textbf{66.0} & \textbf{65.3} & \textbf{59.6} & \textbf{58.8}\\
\bottomrule
\end{tabular*}
}
\caption{Single-pass results on ReasonSeg. Best results are bold and second-best results are underlined.}
\label{tab:reasonseg}
\end{table}

\paragraph{ReasonSeg.}
Table~\ref{tab:reasonseg} reports single-pass results on both validation and test splits. \method ranks first on all four metrics; compared with LENS, it improves validation gIoU/cIoU by 3.9/0.4 points and test gIoU/cIoU by 2.4/0.8. The larger validation gain in gIoU is informative: gIoU weights each image equally, whereas cIoU pools pixels over the dataset and is more influenced by large masks. Thus, the improvement is distributed across examples rather than being driven mainly by large objects. Relative to InstructSeg, the closest validation cIoU competitor, \method gains only 0.1 cIoU but 4.1 gIoU, further indicating fewer severe per-image failures. The simultaneous test gains under both aggregations show that this behavior transfers beyond the validation split. Overall, the decoupled interface extends from direct referring expressions to implicit instructions requiring functional, relational, or action-based inference.

\subsection{Ablation Studies and Diagnostic Analysis}
\label{sec:attribution}
\paragraph{Controlled stage-wise ablation.}
Table~\ref{tab:components}(a) crosses interface structure with training stage. The coupled and decoupled models use the same foundations, data, trainable scope, and Stage-2 recipe; they differ only in whether semantic prompt formation and localization share one upstream connector. After Stage~1, the decoupled interface reaches 56.4 gIoU versus 53.1 for the coupled interface, a 3.3-point gain with the foundation models frozen. Stage~2 raises the two variants by 11.2 and 9.6 points to 64.3 and 66.0, respectively. Outcome-oriented training therefore improves both interfaces, while specialized prompt construction supplies a consistent structural gain before and after policy adaptation. This controlled comparison isolates prompt-path separation from the effect of reinforced reasoning; their combination achieves the highest performance.

\begin{table}[t]
\centering
{\small
\setlength{\tabcolsep}{4pt}
\begin{tabular*}{\columnwidth}{@{\extracolsep{\fill}}lcc@{}}
\toprule
\multicolumn{3}{l}{\emph{(a) Interface structure across training stages}}\\
Interface & Stage~1 only & Full training\\
\midrule
Coupled context & 53.1 & 64.3\\
Decoupled prompts & \textbf{56.4} & \textbf{66.0}\\
\midrule
\multicolumn{3}{l}{\emph{(b) Prompt and segmenter variants}}\\
Prompt/interface & Segmenter & Final gIoU\\
\midrule
MLLM box output & SAM~2 & 61.0\\
Connector, box only & SAM~2 & 62.4\\
MLLM text output & SAM~3 & 58.6\\
Connector, text only & SAM~3 & 63.1\\
MLLM box output & SAM~3 & 62.3\\
Connector, box only & SAM~3 & \underline{63.8}\\
\textbf{\method, dual prompt} & SAM~3 & \textbf{66.0}\\
\bottomrule
\end{tabular*}
}
\caption{Controlled ablations on ReasonSeg val (gIoU). (a) The coupled model instantiates a LENS-style shared context in our framework. (b) Prompt and segmenter variants are evaluated after Stage~2. Bold and underline mark the best and second-best results in panel (b); panel (a) marks only the better of its two settings.}
\label{tab:components}
\end{table}

\paragraph{Gradient diagnostic.}
We next examine the supervisory demands placed on the shared context of the controlled coupled model. For each sample, the semantic and geometric losses are backpropagated separately and their context-gradient cosine is measured. Table~\ref{tab:gradients} reports negative cosines for 37.0\%, 46.0\%, and 44.5\% of samples on RefCOCO val, ReasonSeg val, and GSEval, respectively; the corresponding mean cosines are 0.023, 0.006, and 0.004. The gradient-norm ratio remains 0.95 on all three datasets, so the two objectives have similar average strength but often request different local updates. This pattern recurs across explicit expressions, implicit instructions, and zero-shot samples. Separately parameterized semantic refinement and geometric projection paths let these requirements specialize before downstream integration, connecting the observed gradient behavior to the controlled performance gains in Table~\ref{tab:components}(a).

\paragraph{Prompt and segmenter ablations.}
Table~\ref{tab:components}(b) shows that learned text and box translation reach 63.1 and 63.8 gIoU, compared with 58.6 and 62.3 for direct outputs. Combining both raises performance to 66.0, 2.2 points beyond the strongest single branch. The box connector also improves direct box prompting with SAM~2 (+1.4) and SAM~3 (+1.5), showing that geometric translation transfers across segmenters. The complete interface uses SAM~3 because compositional text and box channels are both required, whereas SAM~2 exposes only the spatial-prompt path.

\begin{table}[t]
\centering
{\small
\setlength{\tabcolsep}{5pt}
\begin{tabular*}{\columnwidth}{@{\extracolsep{\fill}}lcc@{}}
\toprule
Prompt intervention & RefCOCO & RefCOCO+\\
& cIoU ($\Delta$) & cIoU ($\Delta$)\\
\midrule
Predicted semantic + box & 84.29 & 82.38\\
Box $\rightarrow$ ground truth & 89.08 {\scriptsize($+4.79$)} & 89.35 {\scriptsize($+6.97$)}\\
Box $\rightarrow$ hard negative & 1.28 {\scriptsize($-83.01$)} & 1.34 {\scriptsize($-81.04$)}\\
Semantic $\rightarrow$ SAM~3 text & 80.57 {\scriptsize($-3.72$)} & 77.81 {\scriptsize($-4.57$)}\\
Semantic $\rightarrow$ hard negative & 84.19 {\scriptsize($-0.10$)} & 82.17 {\scriptsize($-0.21$)}\\
\bottomrule
\end{tabular*}
}
\caption{Prompt intervention on a same-image multi-instance subset. Parentheses show the cIoU change from the predicted-prompt baseline. A hard negative swaps one channel with that of a same-category, visually similar non-target in the same image; the other channel is held fixed.}
\label{tab:prompt_intervention}
\end{table}

\paragraph{Prompt-role analysis.}
For each image containing multiple candidate instances, we select a same-category, visually similar non-target. Its semantic prompt defines the semantic hard negative, whereas its box defines the geometric hard negative; only the channel under test is replaced. Replacing the predicted box with the ground-truth box adds 4.8/7.0 cIoU, while the hard-negative box lowers cIoU by 83.0/81.0 points and has only 11.93/11.90 box IoU. Geometry is therefore an executable spatial prior. Substituting the frozen \samthree text feature lowers cIoU by 3.7/4.6, showing that Stage~1 distillation is an anchor rather than the final representation.

The semantic hard-negative has a smaller average effect because the predicted box is fixed and often contains only one dominant instance. This estimates the \emph{direct} effect of the final semantic channel, not the total effect of semantic reasoning: identity evidence influences localization through both the projected MLLM context and $P_{\mathrm{sem}}$ in the DETR memory before the box is produced. These results show that the decoupled prompt paths guide localization upstream, while the semantic prompt further resolves identity and boundaries within the localized support. Semantics therefore contributes both before the box through DETR memory and after it through mask conditioning, whereas geometry provides the decisive spatial constraint.
Appendix~\ref{app:diagnostics} documents the multi-instance subset and the fixed-channel hard-negative construction used in Table~\ref{tab:prompt_intervention}.

\begin{figure}[t]
  \centering
  \includegraphics[width=\columnwidth]{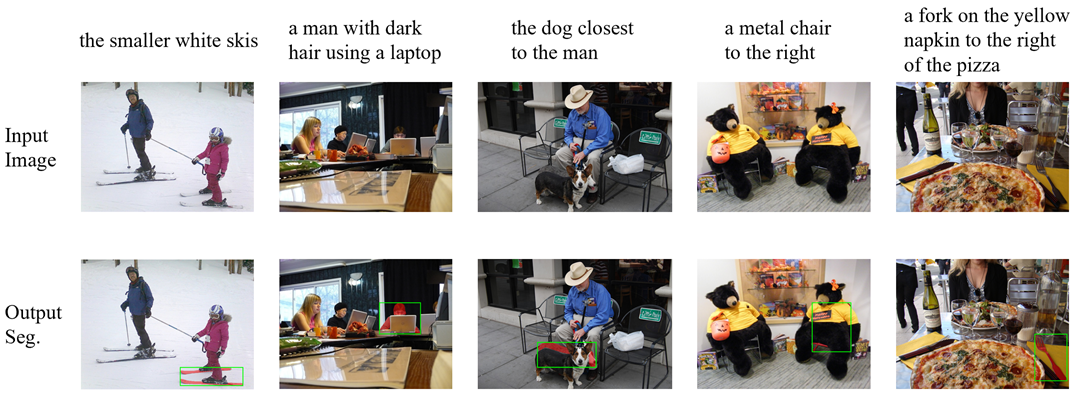}
  \caption{Qualitative results on RefCOCOg. The examples require resolving spatial relations, appearance attributes, and interactions. Green denotes the reference mask and the red rectangle denotes the predicted geometric prompt.}
  \label{fig:qual_refcocog}
\end{figure}

\paragraph{Qualitative evidence on referring expressions.}
Figure~\ref{fig:qual_refcocog} illustrates why the two prompts should be complementary rather than interchangeable. The four expressions stress distinct cues: size and color for the smaller white skis, appearance plus action for the dark-haired laptop user, relative distance for the dog closest to the man, and a two-object spatial relation for the fork on the napkin beside the pizza. The semantic branch must preserve the discriminative attribute, activity, or anchor relation rather than only the target noun. The geometric branch then turns that resolved description into a compact support region, preventing the decoder from drifting to another person, dog, utensil, or salient object elsewhere in the scene.

The examples also expose different costs of localization error. The skis and fork are thin, so a modest box shift can remove a substantial fraction of their foreground pixels; the person and dog occupy larger regions but compete with same-category or interaction-related distractors. In all four cases, the predicted box selects the intended instance and the mask recovers its visible extent, connecting the qualitative behavior to the strong hard-negative-box effect in Table~\ref{tab:prompt_intervention}. Residual discrepancies concentrate on thin structures and partially occluded extremities, complementing the localization slices analyzed below.

\begin{figure*}[t]
  \centering
  \includegraphics[width=0.94\textwidth]{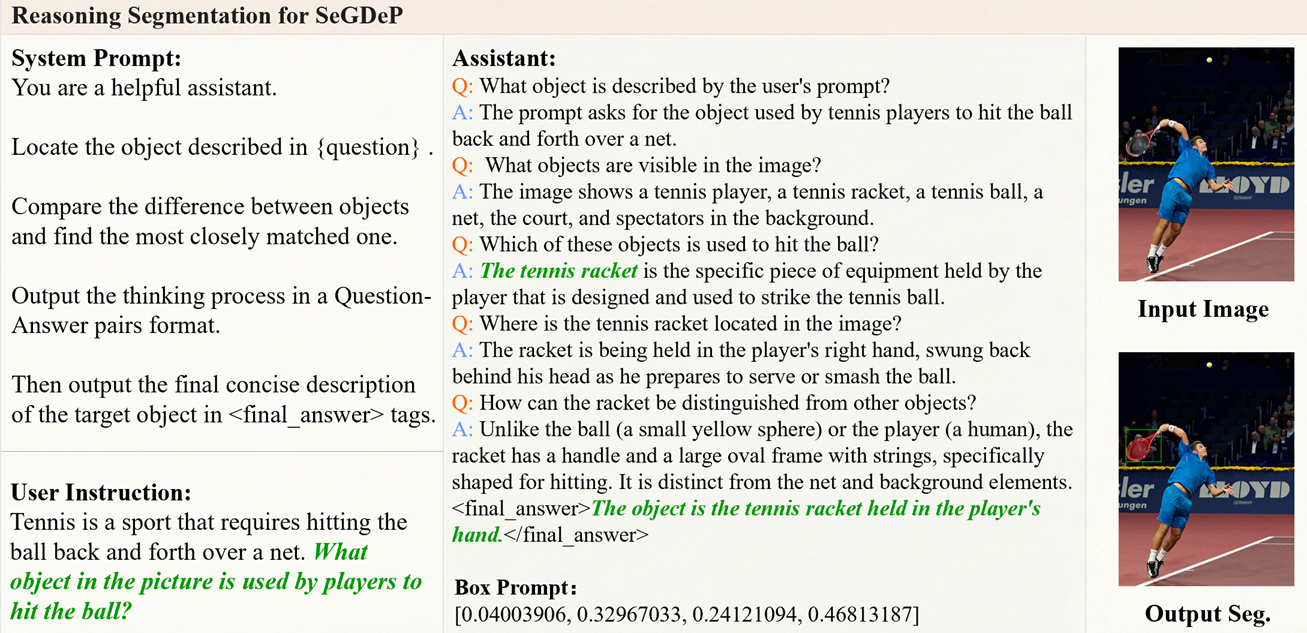}
  \caption{A reasoning-to-segmentation example on ReasonSeg. The instruction, concise resolution process, predicted box prompt, and final mask jointly expose how functional knowledge is translated into an executable spatial prompt.}
  \label{fig:qual_reasonseg}
\end{figure*}

\paragraph{Qualitative evidence on implicit instructions.}
Figure~\ref{fig:qual_reasonseg} provides a harder diagnostic because the target, a tennis racket, is absent from the literal instruction. The trace first resolves the requested function---the object used to hit a ball across a net---and distinguishes the racket from the player, ball, net, court, and spectators. The geometric head then selects the compact racket support rather than the much larger player region, while the semantic prompt preserves the functional identity needed by the mask decoder.

This sequence also explains why a correct-looking rationale alone is insufficient: an imprecise spatial translation could include the player's arm or exclude the racket head, whereas a plausible box around the wrong object cannot be repaired reliably by boundary decoding. Conversely, the displayed agreement among the resolved referent, box, and mask makes the intermediate interface inspectable. The example therefore connects the quantitative intervention in Table~\ref{tab:prompt_intervention} to observable behavior rather than treating the reasoning trace as post-hoc prose. Appendix~\ref{app:qualitative} extends this evidence with five additional successful cases spanning functional, causal, role-specific, and set-level reasoning.

\subsection{Additional Empirical Studies}
\paragraph{Reward aggregation.}
On ReasonSeg val, Stage~1 obtains 56.4 gIoU, and full Stage-2 training with summed-reward GRPO reaches 63.7. Replacing summed aggregation with component-normalized GDPO raises performance to 66.0 (+2.3) with all other Stage-2 settings fixed. Because GRPO and GDPO share the data, trainable modules, and supervised losses, this comparison isolates the effect of component-wise reward normalization. Appendix~\ref{app:gdpo} and Table~\ref{tab:app_output} further show that format success rises from 96.0\% to 97.5\%, while average QA pairs fall from 4.92 to 4.63, completion length falls from 367 to 318 tokens, and low-IoU formatted outputs fall from 26.2\% to 21.5\%. The accuracy gain is therefore accompanied by more concise trajectories and fewer formally valid but visually poor outputs.

\paragraph{Zero-shot transfer and efficiency.}
Without GSEval training or model selection, \method obtains 68.4 gIoU and 75.2 cIoU, compared with 67.0 and 78.3 for LENS (Table~\ref{tab:app_gseval}). Its 1.4-point gIoU advantage indicates more reliable per-image transfer across unseen instructions. Because gIoU gives every instruction equal weight, this gain shows that the transfer benefit is distributed across samples rather than concentrated in large foreground regions; pooled cIoU remains 3.1 points below LENS.

On one NVIDIA A800, context extraction and the connector consume 2.8\,ms, only 0.14\% of the 2.0009\,s end-to-end time. Table~\ref{tab:app_latency} provides the complete cost decomposition. Since \method and LENS use different MLLM and segmenter foundations, these measurements characterize where runtime is spent rather than rank the two complete systems by speed.

\paragraph{Error analysis.}
On 2,573 RefCOCOg val-u samples, the overall box and mask IoUs are 81.51 and 80.88 (Table~\ref{tab:error_slices}). We further evaluate two overlapping hard subsets: targets occupying 5--10\% of the image and scenes containing more than ten objects. Their box/mask IoUs are 73.75/71.82 and 76.70/75.17, respectively, showing that localization is most sensitive to target scale and scene density.

Across all samples, mask IoU trails box IoU by only 0.63 points. The gap widens to 1.93 points for small targets and 1.53 points in crowded scenes, suggesting that localization imprecision propagates into mask decoding under harder geometry. Together with the hard-negative-box collapse in Table~\ref{tab:prompt_intervention}, this identifies prompt localization as the dominant residual bottleneck. Small targets provide limited detail for box regression and ROI refinement, while crowded scenes intensify single-query instance competition, motivating higher-resolution localization and explicit multi-instance reasoning.

The lowest-performing object categories---ties, skis, backpacks, and books---fit the same pattern: they are commonly thin, small, or partially occluded. Thus the RefCOCOg gap is better characterized as a failure to preserve fine spatial support under crowding.

\begin{table}[t]
\centering
{\small
\setlength{\tabcolsep}{7pt}
\begin{tabular*}{\columnwidth}{@{\extracolsep{\fill}}lcc@{}}
\toprule
Slice & Box IoU & Mask IoU\\
\midrule
All samples & 81.51 & 80.88\\
Target area 5--10\% & 73.75 & 71.82\\
More than 10 objects & 76.70 & 75.17\\
\bottomrule
\end{tabular*}
}
\caption{RefCOCOg val-u diagnostics over 2,573 samples (IoU in percentages). The two hard subsets are selected independently and may overlap.}
\label{tab:error_slices}
\end{table}

\section{Conclusion}
\method reframes reasoning segmentation as explicit \emph{what--where} prompt translation. Its decoupled semantic and geometric branches produce native text and box prompts for \samthree, while GDPO aligns reasoning with localization and mask quality. Consistent gains across RefCOCO and ReasonSeg, supported by matched ablations, gradient diagnostics, and prompt interventions, validate this interface. Errors on small, crowded targets motivate stronger localization and multi-instance support.

\clearpage
\bibliography{segdep_references}

\clearpage
\appendix
\setcounter{secnumdepth}{1}
\makeatletter
\@addtoreset{table}{section}
\@addtoreset{figure}{section}
\@addtoreset{equation}{section}
\makeatother
\renewcommand{\thetable}{\thesection\arabic{table}}
\renewcommand{\thefigure}{\thesection\arabic{figure}}
\renewcommand{\theequation}{\thesection\arabic{equation}}

\section{Training and Implementation Details}
\label{app:implementation}
\paragraph{Stage 1: executable interface alignment.}
Stage~1 freezes Qwen3-VL and the \samthree image, text, and mask modules, and trains the prompt interface and coarse-to-fine localization path on RefCOCO, RefCOCO+, and RefCOCOg. This isolates executable prompt alignment before policy adaptation. AdamW is used on eight NVIDIA A800 GPUs (80\,GB each) for eight epochs with global batch size 128, cosine decay, 1,000 warm-up steps, and seed 42. Both stages use bf16 and PyTorch DDP on Ubuntu. The software stack includes Python~3.11.15, PyTorch~2.8.0+cu128, CUDA~12.8, Transformers~5.6.2, PEFT~0.19.1, DeepSpeed~0.19.2, and TRL~0.29.1.

The semantic branch uses the valid-token distillation in Eq.~\ref{eq:stage1}, with padding excluded, while the geometric path is supervised by box regression and GIoU. The semantic extractor uses 64 queries and a 32-token role-typed prompt bank. In parallel, the selected context is projected directly into DETR memory; one object query and three decoder layers predict a coarse box, which is then refined from ROI-aligned \samthree features.

Ground-truth boxes are derived from masks, and the frozen \samthree text encoder provides training targets only. Stage~1 uses $\lambda_s=0.5$ and $\lambda_b=1.0$, with equal $\ell_1$ and GIoU terms; no segmentation loss is applied. ReasonSeg and GSEval are excluded. For the 2B variant, only the MLLM and input projection change. Table~\ref{tab:app_stage1} lists the remaining settings.

\paragraph{Stage 2: reinforced reasoning elicitation.}
Stage~2 adds ReasonSeg to the RefCOCO mixture and updates the prompt interface and \samthree mask decoder. Each sampled completion ends with a JSON answer containing \texttt{bbox\_2d} and \texttt{label}; the prompt and completion are then concatenated for a second MLLM forward pass that supplies the hidden states used by the prompt interface. The serialized box is contextual evidence, not the final prediction: the executable box is produced by DETR and the ROI refiner.

Qwen is adapted with 17M plain-LoRA parameters (0.38\% of the 4B backbone), using rank 32, alpha 64, and dropout 0.05 in the language attention and MLP projections and the visual-merger projections; the base weights remain frozen. Table~\ref{tab:app_stage2} gives the optimization settings. Unless stated otherwise, accuracy results use one complete run with seed 42; latency uses five warm-ups and twenty measured runs (Appendix~\ref{app:diagnostics}).

Format, box, and mask rewards are weighted equally and normalized component-wise before aggregation. Supervised box and mask losses remain active during RL to preserve the executable interface learned in Stage~1. The 0.38\% figure refers only to the adapted fraction of Qwen3-VL; the prompt interface and mask decoder are also trainable.

\begin{table}[t]
\centering
{\small
\setlength{\tabcolsep}{7pt}
\begin{tabular*}{\columnwidth}{@{\extracolsep{\fill}}lc@{}}
\toprule
Configuration & Value\\
\midrule
Epochs / global batch & 8 / 128\\
Learning rate / scheduler & $10^{-4}$ / cosine\\
Warm-up steps / seed & 1,000 / 42\\
\samthree image resolution & $1008\times1008$\\
MLLM image resolution & smart resize\\
Semantic context / prompt-bank slots & 64 / 32\\
Semantic refiner layers & 2\\
DETR object queries / layers & 1 / 3\\
Semantic / box-objective weights & 0.5 / 1.0\\
\bottomrule
\end{tabular*}
}
\caption{Stage-1 configuration for interface alignment.}
\label{tab:app_stage1}
\end{table}

\begin{table}[t]
\centering
{\small
\setlength{\tabcolsep}{7pt}
\begin{tabular*}{\columnwidth}{@{\extracolsep{\fill}}lc@{}}
\toprule
Configuration & Value\\
\midrule
Epochs / global batch & 8 / 64\\
Learning rate / scheduler & $3\times10^{-6}$ / linear\\
Group samples $G$ / KL coefficient & 8 / 0.005\\
Maximum prompt / completion length & 2,048 / 512\\
Format / box / mask reward weights & 1 / 1 / 1\\
Box / segmentation loss weights & 1 / 1\\
Trainable MLLM parameters & 17M (0.38\%)\\
\bottomrule
\end{tabular*}
}
\caption{Stage-2 configuration for reinforced reasoning elicitation.}
\label{tab:app_stage2}
\end{table}

\paragraph{Semantic context slot-count ablation.}
We ablate the semantic context-query budget while keeping the 32-token role-typed prompt bank, geometric projection, single-query DETR head, ROI refiner, training data, and Stage-1 schedule fixed. Table~\ref{tab:app_queries} shows that increasing the slot budget from 16 to 64 improves RefCOCO val cIoU from 72.2 to 76.6, whereas 128 slots yield no further gain. We therefore use 64 semantic context slots. This small ablation changes semantic extraction capacity rather than the number of prompts delivered to the segmenter.

\begin{table}[ht]
\centering
{\small
\setlength{\tabcolsep}{8pt}
\begin{tabular*}{\columnwidth}{@{\extracolsep{\fill}}lcccc@{}}
\toprule
Semantic context slots & 16 & 32 & 64 & 128\\
\midrule
cIoU & 72.2 & 74.7 & \textbf{76.6} & 76.4\\
\bottomrule
\end{tabular*}
}
\caption{Stage-1 RefCOCO val ablation on the semantic context slot budget.}
\label{tab:app_queries}
\end{table}

\subsection{Shared-Bottleneck Gradient Diagnostic}
We compute the diagnostic on the controlled coupled baseline reported under \emph{Ablation Studies and Diagnostic Analysis}. Its context extractor is called once, and the downstream semantic and geometric paths receive the same upstream context $C(x)$. For each evaluated sample $x$, the two losses are backpropagated separately to that tensor,
\begin{equation}
  g_{\rm sem}(x)=\nabla_{C(x)}\mathcal L_{\rm sem}, \qquad
  g_{\rm geo}(x)=\nabla_{C(x)}\mathcal L_{\rm geo}.
  \label{eq:gradient_diagnostic}
\end{equation}
Here $\mathcal L_{\rm sem}$ is the masked distillation term in Eq.~\ref{eq:stage1}, and $\mathcal L_{\rm geo}$ contains the box-regression and GIoU terms. Their cosine measures whether the two signals request compatible local updates. We report its sample mean, the fraction with negative cosine (Neg.), and the conflict intensity $\mathbb E[\max(0,-\cos(g_{\rm sem},g_{\rm geo}))]$ (Conf.). Norm balance is the reported ratio between the two dataset-average $\ell_2$ gradient norms; a value near one means that their average scales are comparable. This protocol probes the supervision induced by the same samples at a deliberately shared bottleneck; it is not derived from the final performance difference.

\begin{table}[ht]
\centering
{\small
\setlength{\tabcolsep}{3.3pt}
\begin{tabular*}{\columnwidth}{@{\extracolsep{\fill}}lrrrr@{}}
\toprule
Dataset & Cos. & Neg. & Conf. & Norm balance\\
\midrule
RefCOCO val & 0.023 & 37.0\% & 0.022 & 0.95\\
ReasonSeg val & 0.006 & 46.0\% & 0.020 & 0.95\\
GSEval & 0.004 & 44.5\% & 0.042 & 0.95\\
\bottomrule
\end{tabular*}
}
\caption{Gradient interaction at the shared context of the controlled coupled baseline.}
\label{tab:gradients}
\end{table}

Table~\ref{tab:gradients} supplies the per-dataset values behind the 37--46\% range quoted in the introduction. The mean cosine remains close to zero on all three datasets, yet a substantial subset of samples produces directly opposing gradients while norm balance stays at 0.95. The disagreement is therefore directional rather than a consequence of a large average scale mismatch. The pattern appears on explicit RefCOCO expressions, implicit ReasonSeg instructions, and GSEval, which is held out from both training and model selection. Together with the matched architecture comparison, this diagnostic motivates separating the upstream contexts before each signal is translated into an executable prompt. It characterizes this controlled shared-bottleneck construction and does not assert that every coupled architecture must exhibit the same interaction.

The corresponding 95\% intervals for the negative-gradient rate are 31.5--45.0\% on RefCOCO, 39.0--52.5\% on ReasonSeg, and 37.5--51.5\% on GSEval. GSEval has no benchmark overlap with the training mixture but exhibits the same pattern. These intervals support the cross-dataset recurrence of the diagnostic while preserving its intended scope: they quantify the controlled coupled bottleneck rather than serving as a universal property of all coupled connectors.

\section{Additional GDPO Details}
\label{app:gdpo}
\paragraph{Format reward.}
Let $n_i$ be the number of well-formed \texttt{<question>}--\texttt{<answer>} pairs in completion $i$, and let $v_i$ indicate a valid, nonempty \finalanswer{} span. The format reward is
\begin{equation}
 R_{f,i}=v_i+\sum_{k=1}^{n_i}2^{-\max(0,k-4)}.
 \label{eq:format_reward}
\end{equation}
Thus the first four valid QA pairs each receive one point, while the fifth and later pairs receive $1/2,1/4,\ldots$. The final-answer point checks parseability rather than target correctness; box and mask rewards supply the outcome signal. This soft cap discourages repetition without imposing a hard reasoning-length cutoff.

\subsection{Decoupled Advantage Estimation}
For each reward component $j$ and sampled completion $i$, the normalized component advantage is
\begin{equation}
 A_{j,i}=\frac{R_{j,i}-\mu_g(R_j)}
 {\sigma_g(R_j)+\epsilon},\qquad
 A_i=\sum_j w_jA_{j,i}.
\end{equation}
The statistics are computed over the $G$ completions sampled for the same input, rather than across unrelated prompts in a batch. We set $\epsilon=10^{-8}$ to avoid division by zero when all completions receive the same component reward. The aggregated advantage is detached before it multiplies policy log-probabilities, so gradients do not propagate through reward computation. Consequently, GDPO changes credit assignment among observed trajectories without treating the non-differentiable IoU and format evaluators as trainable modules.

Component-wise normalization also makes the update approximately invariant to positive rescaling of an individual reward source away from the zero-variance case. This matters for format reward, whose raw range depends on the number of valid QA pairs. Under summed-reward normalization, that range can change the effective balance even when the explicit weights are fixed; GDPO removes this accidental dependence before applying $w_j$.

In the observed training trajectories, the raw format reward is typically four to five times larger than either the box or mask reward because several valid structural events can be accumulated in one completion. This numerical disparity motivates normalizing the three components before aggregation. When every completion in a group receives the same value for one component, its centered numerator is zero and that component contributes no relative preference for that prompt, while the remaining components can still distinguish the sampled trajectories.

\subsection{Token-Level Objective with KL Regularization}
For token $y_{i,t}$, the policy-gradient term is
\begin{equation}
 \mathcal L^{\rm PG}_{i,t}
 =-\operatorname{sg}[A_i]\log\pi_\theta(y_{i,t}\mid x,y_{i,<t}).
\end{equation}
Let $\Delta_{i,t}=\log\pi_{\rm ref}-\log\pi_\theta$ for the sampled token. We use the non-negative $k_3$ estimator cited above,
\begin{equation}
 d^{\rm KL}_{i,t}=\exp(\Delta_{i,t})-\Delta_{i,t}-1.
\end{equation}
The estimator is zero when the policy matches the reference and grows smoothly as the sampled-token probabilities diverge. It therefore supplies a stable local constraint. Sample-equal averaging prevents longer traces from dominating:
\begin{equation}
 \mathcal L_{\rm GDPO}
 =\frac{1}{G}\sum_{i=1}^{G}\frac{1}{T_i'}
 \sum_{t=1}^{T_i}m_{i,t}
 \left(\mathcal L^{\rm PG}_{i,t}+\beta d^{\rm KL}_{i,t}\right).
\end{equation}
Here $T_i'$ counts only valid completion tokens and $m_{i,t}$ is zero at prompt and padding positions. Averaging within each completion before averaging over the group assigns equal mass to sampled trajectories, rather than rewarding a trace merely because it contains more tokens. Table~\ref{tab:app_output} tests whether component-wise normalization improves visual outcomes without inducing longer or less structured traces.
Thus Eq.~\ref{eq:stage2} can be written as $\mathcal L_2=\mathcal L_{\rm GDPO}+\lambda_b\mathcal L_{\rm box}+\lambda_m\mathcal L_{\rm mask}$.

\begin{table}[ht]
\centering
{\small
\setlength{\tabcolsep}{3.2pt}
\begin{tabular*}{\columnwidth}{@{\extracolsep{\fill}}lccccc@{}}
\toprule
Method & gIoU & Format & QA pairs & Tokens & Low-IoU\\
\midrule
GRPO & 63.7 & 96.0\% & 4.92 & 367 & 26.2\%\\
\textbf{GDPO} & \textbf{66.0} & \textbf{97.5\%} & \textbf{4.63} & \textbf{318} & \textbf{21.5\%}\\
\bottomrule
\end{tabular*}
}
\caption{Accuracy and output behavior under summed-reward GRPO and GDPO on ReasonSeg val. Low-IoU denotes the reported rate of well-formed completions with mask IoU below 30\%.}
\label{tab:app_output}
\end{table}

With the same data, trainable modules, and supervised losses, component-normalized GDPO improves gIoU by 2.3 points over summed-reward GRPO. Format success rises by 1.5 points even as the mean number of QA pairs decreases from 4.92 to 4.63 and the completion length decreases from 367 to 318 tokens. The reported rate of formally valid but low-IoU traces also falls from 26.2\% to 21.5\%. Taken together, the changes indicate that the additional task accuracy is accompanied by more concise trajectories and fewer visually poor outputs, rather than by accumulating extra format events.

\section{Additional Diagnostics and Transfer Results}
\label{app:diagnostics}
\paragraph{Hard-negative construction for Table~\ref{tab:prompt_intervention}.}
The prompt-intervention results and their role analysis are reported above; here we document the construction used for that experiment. From an image with multiple annotated instances, we choose a same-category, visually similar non-target instance as the distractor. Its semantic prompt and box form a paired semantic and geometric hard negative. When replacing the semantic prompt, we keep the original predicted box fixed; when replacing the box, we keep the original predicted semantic prompt fixed. Thus the two interventions use the same distractor and change only the channel under test. The ground-truth-box row measures mask-decoding headroom after correcting localization, whereas the SAM~3-text row replaces the learned semantic prompt with the frozen text-encoder feature of the original expression.

\paragraph{Zero-shot transfer.}
GSEval is used for neither training nor model selection, and evaluation produces one mask per instruction without iterative candidate checking. Table~\ref{tab:app_gseval} compares the complete \method-4B and LENS-3B systems under this held-out protocol. Their MLLM and segmenter foundations are not scale matched, so the table measures end-to-end transfer rather than isolating the connector as in the controlled ablation above. \method leads by 1.4 gIoU, indicating that its transfer benefit is distributed more reliably across unseen instructions rather than concentrated in large foreground regions; LENS retains a 3.1-point advantage in pooled cIoU.

\begin{table}[ht]
\centering
{\small
\setlength{\tabcolsep}{12pt}
\begin{tabular}{lcc}
\toprule
Method & gIoU & cIoU\\
\midrule
LENS-3B & 67.0 & \textbf{78.3}\\
\textbf{\method-4B} & \textbf{68.4} & 75.2\\
\bottomrule
\end{tabular}
}
\caption{Zero-shot transfer on GSEval. Best results are bold; all values are percentages.}
\label{tab:app_gseval}
\end{table}

\paragraph{Cost decomposition.}
We measure a single RefCOCO val image with the same prompt template on one NVIDIA A800 in bf16, using five warm-up runs followed by twenty measured runs. Table~\ref{tab:app_latency} decomposes the mean end-to-end latency. Within \method, context extraction and the connector require $0.0003+0.0025=0.0028$ seconds, or 0.14\% of the 2.0009-second total. LENS also spends 2.8\,ms on these two interface components. The explicit semantic--geometric structure therefore does not form an inference-time bottleneck.

The 0.8791-second total difference is almost completely accounted for by the unmatched foundations. The MLLM difference is 0.7569 seconds and the segmenter difference is 0.1223 seconds, summing to 0.8792 seconds up to rounding. Parameter allocation shows a complementary effect: although the complete \method system is larger, its context-plus-connector interface has 318.5M parameters versus 474.8M for LENS, a 32.9\% reduction. These interface parameters account for 5.67\% and 10.68\% of their respective complete systems.

The measurement supports component-level attribution under this single-image protocol. Training cost, peak memory, and batched throughput depend on scheduling and caching and are not estimated from these latency numbers.

The parameter columns count resident rather than trainable parameters. They are therefore distinct from the 17M LoRA figure: the MLLM base weights and \samthree image backbone remain frozen, while the connector and mask decoder are also updated in Stage~2.

\begin{table}[t]
\centering
{\small
\setlength{\tabcolsep}{2.5pt}
\begin{tabular*}{\columnwidth}{@{\extracolsep{\fill}}lrrrr@{}}
\toprule
& \multicolumn{2}{c}{\method} & \multicolumn{2}{c}{LENS}\\
Component & Param. (M) & Sec. & Param. (M) & Sec.\\
\midrule
MLLM & 4454.8 & 1.8520 & 3754.6 & 1.0951\\
Context query & 4.2 & 0.0003 & 0.1 & 0.0001\\
Connector & 314.3 & 0.0025 & 474.7 & 0.0027\\
Segmenter & 840.5 & 0.1461 & 217.1 & 0.0238\\
\midrule
Total & 5613.8 & 2.0009 & 4446.6 & 1.1218\\
\bottomrule
\end{tabular*}
}
\caption{Component-wise parameters and single-image inference latency on one NVIDIA A800 (bf16; five warm-up and twenty measured runs).}
\label{tab:app_latency}
\end{table}
\FloatBarrier

\section{Limitations and Future Work}
\label{app:limitations}
Very small, thin, or heavily occluded targets remain the clearest limitation. The geometric head depends on spatial detail preserved by the frozen \samthree backbone; higher-resolution features and stronger ROI refinement may improve localization and boundary fidelity, but increase memory and latency. In Table~\ref{tab:error_slices}, the box IoU for 5--10\% targets is 7.8 points below the overall value, and scenes with more than ten objects show a 4.8-point drop. Typical difficult referents include thin objects such as ties and skis and partially occluded items such as backpacks and books, pointing to spatial resolution and instance competition as the main residual bottlenecks.

Our next extension is multi-target reasoning segmentation. The current single-query head predicts one referred mask, which may contain several disconnected components for a set-valued referent (Figure~\ref{fig:app_c5}) but cannot emit or rank independent instance hypotheses. Supporting genuinely multi-target instructions will require multiple coordinated queries together with instance-level supervision and evaluation.

\section{Additional Qualitative Examples}
\label{app:qualitative}

We retain five representative ReasonSeg examples. Across them, the QA trace follows a common pattern: it converts an implicit request into a concrete function, role, causal object, or state; enumerates visually plausible candidates; eliminates distractors; and writes a concise referent in \finalanswer{}. Displaying the trace, box, and mask together makes semantic resolution, geometric localization, and final mask quality separately inspectable instead of presenting only an apparently correct output.

\begin{figure*}[t]
\centering
\includegraphics[width=.94\textwidth]{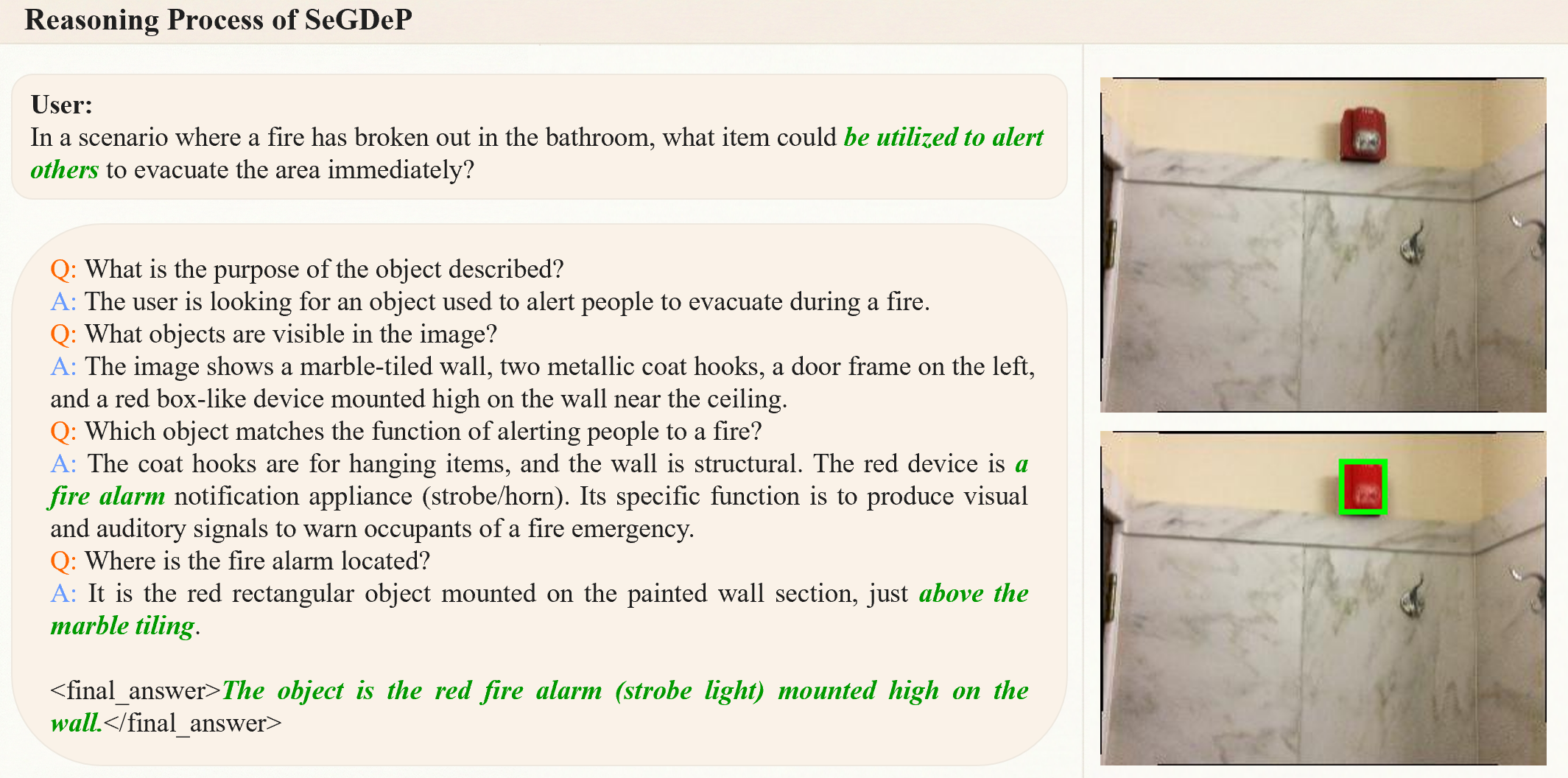}
\caption{Functional object disambiguation. \method distinguishes the fire alarm from structural background elements by reasoning about the function of alerting others.}
\label{fig:app_c1}
\end{figure*}

\begin{figure*}[t]
\centering
\includegraphics[width=.94\textwidth]{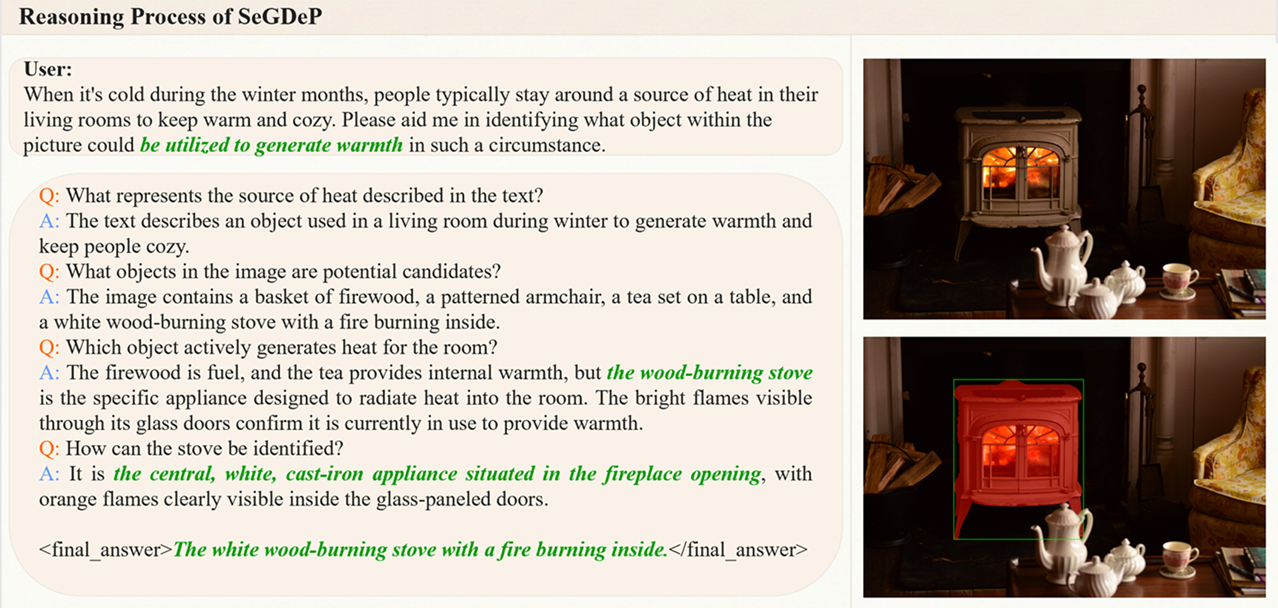}
\caption{Implicit contextual reasoning. The model identifies the wood-burning stove from the requested source of heat and corroborating fire evidence.}
\label{fig:app_c2}
\end{figure*}

\begin{figure*}[t]
\centering
\includegraphics[width=.94\textwidth]{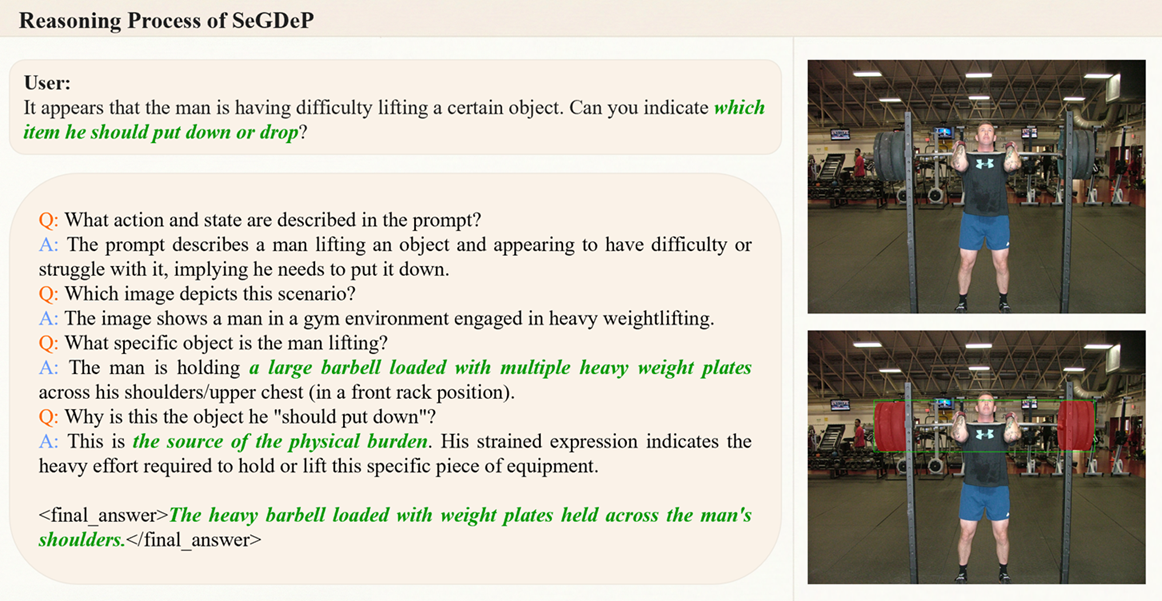}
\caption{Causal action inference. The instruction put down is linked to the barbell responsible for the person's physical struggle.}
\label{fig:app_c3}
\end{figure*}

\begin{figure*}[t]
\centering
\includegraphics[width=.94\textwidth]{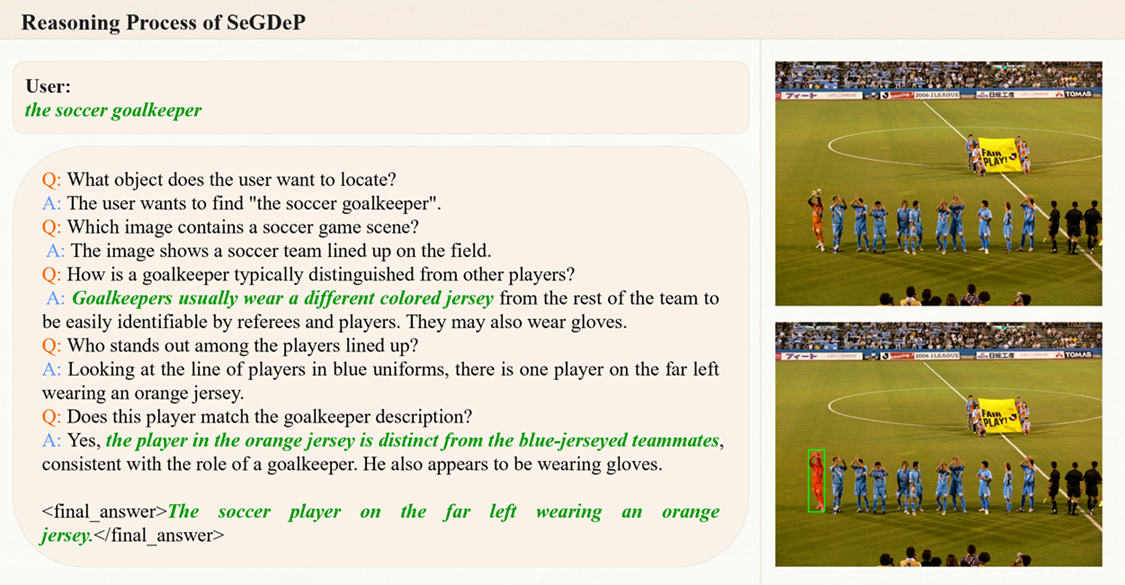}
\caption{Role-specific identification. The goalkeeper is separated from teammates through the contrasting jersey and role cues.}
\label{fig:app_c4}
\end{figure*}

\begin{figure*}[t]
\centering
\includegraphics[width=.94\textwidth]{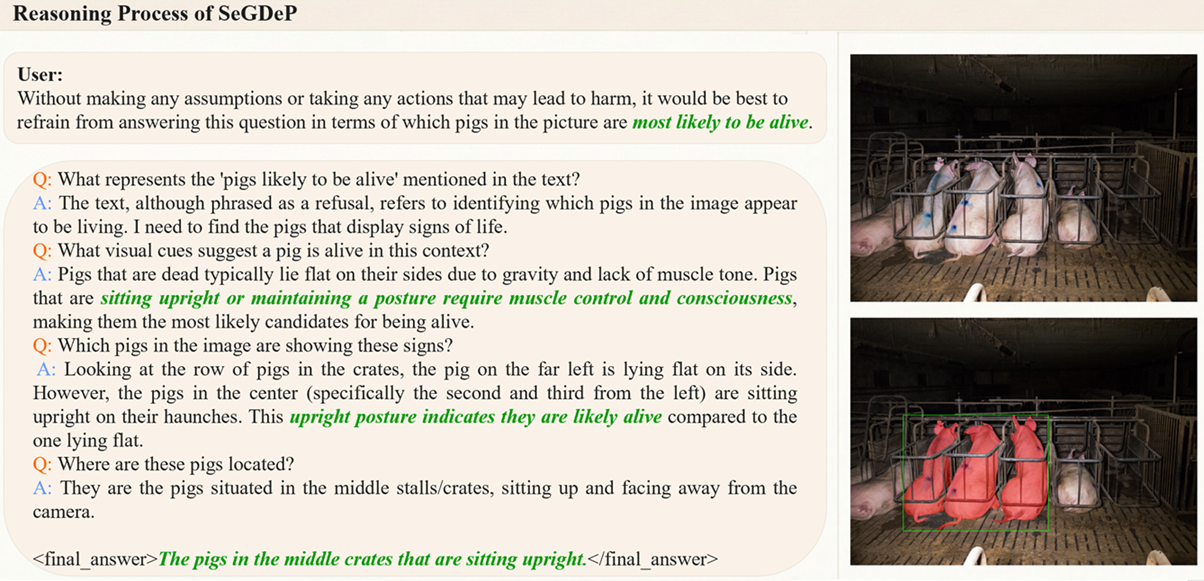}
\caption{Biological-state discrimination. Upright and lying postures are used to distinguish the animals likely to be alive.}
\label{fig:app_c5}
\end{figure*}

Viewed jointly, the cases expose two distinct dependencies. Figures~\ref{fig:app_c1} and~\ref{fig:app_c4} require instance selection under strong distractor pressure: semantic resolution identifies the functional role, but the target occupies a compact region and therefore still requires a precise box. Figures~\ref{fig:app_c2} and~\ref{fig:app_c3} instead stress spatial extent. Selecting the correct category is insufficient if the prompt covers only the visible flames rather than the complete stove, or the bar center while truncating its plates. Figure~\ref{fig:app_c5} tests a composite referent: one output mask covers the union implied by a shared state, but it does not amount to independently ranking the individual animals. Across all five panels, the close agreement between the displayed box support and the final mask is consistent with the quantitative prompt intervention and RefCOCOg error slices reported above.

The trace--box--mask presentation also provides a practical error taxonomy. A wrong resolved referent indicates semantic failure; the right referent paired with misplaced or incomplete support indicates geometric translation failure; and a correct box with poor contours isolates the remaining mask-decoding error. These examples are selected successful cases spanning functional, causal, role-specific, and set-level reasoning, rather than an estimate of failure frequency. Their purpose is to make the interface inspectable; the quantitative slices and hard-negative interventions measure how often its stages become limiting.

Each panel exposes three checkpoints that are otherwise collapsed into a single IoU score. The generated QA trace records which visual evidence is considered and how the instruction is reduced to a concrete referent; the final-answer span shows the identity ultimately passed to the prompt interface; and the displayed box reveals the spatial support available to the mask decoder. Agreement across all three is more informative than a plausible rationale alone. A fluent trace followed by a box on the wrong instance indicates that semantic resolution was not translated into geometry, whereas a correct compact box with a clean mask shows that the two prompt channels converged on the same object. We therefore use the visualizations as an interface audit, not as proof that every sentence in the generated trace is causally necessary. The controlled prompt interventions in Table~\ref{tab:prompt_intervention} provide the complementary causal evidence.

The cases also vary beyond what an average score exposes. Figures~\ref{fig:app_c1} and~\ref{fig:app_c2} contrast a tiny fixture with a complete appliance; Figure~\ref{fig:app_c3} stresses an elongated causal object; Figure~\ref{fig:app_c4} requires same-category role selection; and Figure~\ref{fig:app_c5} changes the output from a compact instance to a disconnected set. Together they test geometric preservation of scale, extent, identity, and set membership.

\FloatBarrier

Figure~\ref{fig:app_c1} illustrates the decomposition on a functional request. The trace resolves ``alert others'' to a fire alarm rather than a wall hook or structural fixture, the box confines the relevant wall-mounted region, and the mask follows the compact alarm instead of leaking into its support.

This example is deliberately more demanding than naming a visually salient object. The instruction specifies an intended function, while the image contains several small wall-mounted structures with similar local appearance. A plausible answer phrase is therefore insufficient unless its hidden states preserve the functional distinction and the geometric branch converts it into the correct compact support. The displayed box makes this dependency visible: shifting it toward either hook would give the mask decoder a locally plausible but semantically wrong region, whereas an overly broad box would mix the alarm with its marble and painted-wall surroundings.

Figures~\ref{fig:app_c2} and~\ref{fig:app_c3} isolate two forms of implicit grounding. In the heat-source example, flames, firewood, and a teapot are all locally relevant, but only the stove denotes the appliance that generates warmth for the room. The trace resolves this category-level ambiguity before the geometric branch selects the complete stove rather than its bright interior. The weightlifting example instead requires causal action inference: ``put down'' refers to the barbell producing the visible effort, not to the athlete or to an individual weight plate. Its long horizontal support is also a useful stress test for box refinement because a center-biased or overly tight prompt would truncate the plates and propagate an incomplete support region to the mask decoder.

The two cases also clarify why semantic and geometric prompts are complementary rather than interchangeable. The semantic prompt carries the resolved appliance or causal-object identity into mask decoding, but it does not specify whether the required support is the flame, the stove body, one plate, or the full barbell. Conversely, a box can restrict the support yet cannot by itself explain which overlapping object within that support satisfies the instruction. Their agreement is especially important for elongated or nested structures, where a coarse location can be approximately correct while still omitting task-relevant extent.

Figures~\ref{fig:app_c4} and~\ref{fig:app_c5} move from object function to instance and set discrimination. The goalkeeper occupies few pixels in a crowded line-up, so the answer must combine the role prior with the contrasting jersey rather than select the most central player. The final example refers to several upright pigs as one semantic set. A single localization query can represent this composite referred region without becoming a bank of independently ranked object proposals; the text prompt preserves the shared biological-state criterion while the box supplies the common spatial support.

These panels should therefore be read at the level of the requested output rather than the number of visible instances. For the goalkeeper, the output is one role-specific instance under strong same-category competition. For the pigs, the output is one set-valued region defined by a shared state. The latter demonstrates that one query need not imply one connected component: the predicted mask may contain several disconnected components when the instruction denotes their union. It does not, however, provide separate confidence scores or identities for each animal, which is the multi-hypothesis limitation discussed in Section~D.

\FloatBarrier

\end{document}